\documentclass[pdflatex,sn-mathphys-num]{sn-jnl}

\usepackage{graphicx}%
\usepackage{multirow}%
\usepackage{amsmath,amssymb,amsfonts}%
\usepackage{amsthm}%
\usepackage{mathrsfs}%
\usepackage[title]{appendix}%
\usepackage{xcolor}%
\usepackage{textcomp}%
\usepackage{manyfoot}%
\usepackage{booktabs}%
\usepackage{algorithm}%
\usepackage{algorithmicx}%
\usepackage{algpseudocode}%
\usepackage{listings}%
\usepackage{subcaption}
\usepackage{float}
\usepackage{caption}
\usepackage{array}
\usepackage{booktabs}
\usepackage{graphicx} 
\usepackage{xcolor}
\usepackage{url}
 \usepackage{booktabs}
\usepackage{array}
\usepackage{makecell} 
\usepackage{graphicx}
\usepackage{caption}
\usepackage{multicol}
\usepackage{verbatim}
\usepackage{hyperref}
\usepackage{float}
\usepackage{listings}
\usepackage{subcaption}
 \usepackage{multirow}
 \usepackage{subcaption}
 \usepackage{longtable}
 \usepackage{amssymb}
\usepackage{listings}
\usepackage{amsmath}
\usepackage{color}
\usepackage{float} 

\definecolor{codegreen}{rgb}{0,0.6,0}
\definecolor{codegray}{rgb}{0.5,0.5,0.5}
\definecolor{codepurple}{rgb}{0.58,0,0.82}
\definecolor{backcolour}{rgb}{0.95,0.95,0.92}

\lstdefinestyle{mystyle}{
    backgroundcolor=\color{backcolour},   
    commentstyle=\color{codegreen},
    keywordstyle=\color{magenta},
    numberstyle=\tiny\color{codegray},
    stringstyle=\color{codepurple},
    basicstyle=\footnotesize,
    breakatwhitespace=false,         
    breaklines=true,                 
    captionpos=b,                    
    keepspaces=true,                 
    numbers=left,                    
    numbersep=5pt,                  
    showspaces=false,                
    showstringspaces=false,
    showtabs=false,                  
    tabsize=2
}

\usepackage{tikz}
\usetikzlibrary{arrows.meta, positioning}

\theoremstyle{thmstyleone}%
\theoremstyle{thmstyletwo}%

\theoremstyle{thmstylethree}%

\begin{document}

\title[Article Title]{Fake News Detection: Comparative Evaluation of BERT-like Models and Large Language Models with Generative AI-Annotated Data}


\author*[1]{\fnm{Shaina} \sur{Raza}}\email{shaina.raza@torontomu.ca}

\author[2]{\fnm{Drai} \sur{Paulen-Patterson}}\email{dpaulen@torontomu.ca}

\author[2]{\fnm{Chen} \sur{Ding}}\email{cding@torontomu.ca}

\affil*[1]{\orgdiv{AI Engineering}, \orgname{Vector Institute}, \orgaddress{\street{College Street}, \city{Toronto}, \postcode{M5G 0C6.}, \state{ON}, \country{Canada}}}

\affil[2]{\orgdiv{Department of Computer Science}, \orgname{Toronto Metropolitan University}, \orgaddress{\street{350 Victoria St}, \city{Toronto}, \postcode{M5B 2K3}, \state{ON}, \country{Canada}}}


\abstract{Fake news poses a significant threat to public opinion and social stability in modern society. This study presents a comparative evaluation of BERT-like encoder-only models and autoregressive decoder-only large language models (LLMs) for fake news detection. We introduce a dataset of news articles labeled with GPT-4 assistance (an AI-labeling method) and verified by human experts to ensure reliability. Both BERT-like encoder-only models and LLMs were fine-tuned on this dataset. Additionally, we developed an instruction-tuned LLM approach with majority voting during inference for label generation. Our analysis reveals that BERT-like models generally outperform LLMs in classification tasks, while LLMs demonstrate superior robustness against text perturbations. Compared to weak labels (distant supervision) data, the results show that AI labels with human supervision achieve better classification results. This study highlights the effectiveness of combining AI-based annotation with human oversight and demonstrates the performance of different families of machine learning models for fake news detection.}




\maketitle

\section{Introduction}
Fake news is defined as false or misleading information intended to deceive or manipulate \cite{raza_fake_2022}. It spreads through various media, including print and social platforms. Fabricated stories, distorted facts, sensationalized headlines, and selectively edited content all fall under the category of fake news \cite{truicua2024danes}. The motives behind its spread range from financial gain to advancing particular agendas that influence public opinion and sow confusion. Significant societal impacts of fake news are evident in events such as the December 2016 incident at Comet Ping Pong in Washington, D.C., where a violent act was incited by a false conspiracy theory\footnote{\url{https://www.marubeni.com/en/research/potomac/backnumber/19.html}}. Additionally, during the 2016 U.S. presidential campaign, manipulated visual evidence was used to spread fake news about Hillary Clinton’s health to influence voter behavior\footnote{\url{https://libguides.lib.cwu.edu/c.php?g=625394&p=4391900}}.  Widespread deepfakes and false news are also influencing voter behavior in the 2024 U.S. Election. 

Research indicates that altering people political opinions is challenging \cite{adammisinformation}; however, strategies aimed at nudging their behaviors are more achievable and are currently being explored. These efforts combine collaborative strategies in journalism with technologies, where artificial intelligence (AI) can play an important role. Advances in machine learning (ML) and deep learning (DL) are also seen in the classification of fake news \cite{sharma2019combating,hamed_review_2023} in the last decade. The development of transformer-based models, such as BERT-like models \cite{kaliyar_fakebert_2021,shu2020fakenewsnet,raza_fake_2022}, has further advanced this research. The  recent advent of generative  (gen.) AI, particularly large language models (LLMs) provides advanced methods for mitigating misinformation \cite{wang_evaluating_2023}. However, these models also paradoxically facilitate the spread of misinformation by generating plausible but false content \cite{longpre2023pretrainer}. 

A primary challenge in using ML-based methods for fake news detection is the availability of accurate and reliable labeling of data. Related works in this field is seen on the datasets such as LIAR \cite{wang2017liar}, FakeNewsNet \cite{shu2020fakenewsnet}, Snopes \cite{vo-lee-2020-facts}, CREDBANK \cite{mitra2015credbank}, the COVID-19 Fake News Dataset\footnote{\url{https://competitions.codalab.org/competitions/26655}}, Fakeddit \cite{nakamura2019r}, MM-COVID \cite{li2020mm}, and NELA-GT \cite{gruppi2020nelagt2019}. These datasets are either labeled through crowdsourcing \cite{mitra2015credbank}, or are available in small samples due to proprietary reasons \cite{shu2020fakenewsnet}. Additionally, utilizing distant labeling \cite{mintz2009distant}, which leverages source-level information from fact-checking websites to label news articles, is a common practice in fake news detection research. This approach is evident in datasets such as NELA-GT \cite{gruppi2020nelagt2019} and Fakeddit \cite{nakamura2019r}. These distant labeling methods are robust but can still introduce source-level biases (e.g., news source they are obtained from) and may not capture the nuanced context of the news articles. 

Experts' annotations typically offer high accuracy but are costly and not scalable for production environments, whereas crowd-sourced methods are labor-intensive and often suffer from inconsistent quality control. Recent explorations into using gen. AI for labeling \cite{gilardi2023chatgpt,tan2024large} have demonstrated potential in accelerating annotation efforts with improved accuracy. However, without human oversight, this approach carries the risk of introducing biases inherent in the AI itself during the labeling process \cite{he2023annollm}.

In this work, we introduce a fake news detection  approach that combines an automated, but human-supervised, annotation process with DL and gen. AI techniques to identify fake news. Our goal is to raise societal awareness of fake news through the use of AI detection methods with its responsible use that are robust and capable of processing large volumes of data. Following the annotation process, we establish baselines by comparing two leading families of models: BERT-like encoder-only models, which excel in classification tasks, and autoregressive decoder-only LLMs such as Mistral-7B \cite{jiang2023mistral} and Llama2-7B \cite{Llama}, which are recognized for their strong performance across a range of NLP tasks. This comparison aims to determine which model family offers superior classification capabilities and under what conditions one may outperform the other.


\subsection*{Contributions}

The specific contributions of this work are:

\begin{itemize}
    \item We release a dataset consisting of 10,000 news articles curated from various sources, collected between May and October 2023, and labeled for fake news detection using a state-of-the-art LLM, such as OpenAI GPT-4 \cite{openai2023}. We utilized prompts with two demonstrations (examples) for binary labeling. We verified the gen. AI labels through human expert reviews to ensure accuracy and reliability.
    \item We conduct a comparative analysis of the performance of BERT-like models and LLMs for the task of fake news detection. BERT-like models typically use a softmax layer over the transformer output to generate probabilistic outputs for classification tasks. In contrast, LLMs, which are not inherently designed for classification, require an adapted approach. We implement a label extraction process that involves computing the model confidence score for each label. This process is repeated multiple times to determine the final classification outcome.
    \item We evaluate the performance of our classifiers using an AI- and human-supervised annotated dataset, alongside another dataset with weak supervision (distant / source-level labeling). This comparison helps determine which setting yields better classification results.
\end{itemize}
Our empirical analysis demonstrates that an AI- and human-supervised labels are more accurate than those obtained through source-level distant supervision for this task of fake news classification (Table \ref{tab:models_performance}). Our findings reveal that while BERT-like models generally excel as classifiers, LLMs demonstrate superior robustness and maintain performance more effectively under perturbations or adversarial testing (Table \ref{tab:model_impact}).


\section{Related Works}
\textbf{Fake News}  
Fake news can be broadly defined as fabricated information that resembles legitimate news content but lacks the editorial standards and processes necessary to ensure accuracy and credibility\cite{tufchi2023comprehensive}. 
The problem of detecting fake news is an important branch of text classification \cite{liu_two-stage_2019}, which focuses on distinguishing real news from fake content. The term “fake news” refers to any misinformation or deceptive content presented as legitimate news, often with the intent to mislead the audience \cite{lu_fake_2022,zhou_survey_2020,allcott2017social,raza_fake_2022}. This includes various forms such as disinformation, which is intentionally false, misinformation, which may be unintentional, and other categories like hoaxes, satires, and clickbait as described in \cite{zhou_survey_2020}. 

\textbf{Deep Learning based Models and Transformer Architecture}
Recent advancements have significantly leveraged ML and DL to improve the accuracy and speed of fake news detection\cite{lu_fake_2022}. For instance, some studies \cite{lu_fake_2022} demonstrate the effectiveness of DL in enhancing the capabilities of fake news classifiers. Other studies \cite{arora_reviewing_2023}, \cite{bonny_detecting_2022,raza_automatic_2021}  show the benefits of countering misinformation through AI, but they also point to persistent challenges such as dataset quality, feature extraction, and the integration of diverse data types \cite{hamed_review_2023}.

Related works \cite{szczepanski2021new} show that transformer- based models like BERT-based models have demonstrated strong performance in fake news detection. The evolution of language styles \cite{raza2019news}, the role of visual elements \cite{qi2019exploiting}, and critical contextual details \cite{fakeNewsData} also play a significant role in enhancing fake news detection accuracy. Some approaches leverage transformer models to analyze both the news content and its social context \cite{raza_fake_2022, fakeNewsData} for a more comprehensive understanding of misinformation patterns. 

Multi-platform and multi-lingual challenges have been addressed to detect fake news across varied contexts \cite{faustini2020fake}, while credibility assessment through ML has demonstrated its role in verification \cite{credibilityBased}. Sentiment analysis techniques leverage emotional tone to infer falsity \cite{SentimentAnalysis,truica2017classification}, and dual transformer models combine content and social context for enhanced detection \cite{raza_fake_2022}. Multi-modal integration, incorporating text, images, and publisher features, has shown improved performance in social media contexts \cite{evaluatingEffect}. Hybrid models, such as transformers model with traditional ML methods also optimize classification precision and adaptability \cite{HybridBert,olan2024fake}. Semantic relationships captured through embeddings \cite{ilie2021context} and transformer-based models, such as BERT and GPT are shown to facilitate processing of long text sequences \cite{truicua2022misrobaerta}. Techniques like sentence and document embeddings \cite{truicua2022awakened,truicua2023s}, ensemble deep neural networks \cite{truicua2024danes}, and real-time misinformation detection algorithms \cite{jain2016towards} show better detection methods. Beyond detection, strategies for social network immunization \cite{petrescu2021sparse} and community-targeted interventions \cite{zhao2021detecting} offer effective tools for mitigating the spread of misinformation.

Recent advancements in fake news detection also emphasize leveraging various embedding techniques and transformer models. For example, word embeddings \cite{ilie2021context} have been used to capture semantic relationships, while transformer-based models like BERT and GPT have shown remarkable accuracy in processing long and complex texts \cite{truicua2022misrobaerta}. Sentence transformers \cite{truicua2022awakened}, and document embeddings \cite{truicua2023s} further enhance detection by capturing deeper contextual nuances at both sentence and document levels. Additionally, studies have highlighted the integration of network information \cite{truicua2024danes}, such as social graph features, to improve the robustness and accuracy of fake news detection. 

\textbf{Knowledge Graphs} 
Some other work has focused on incorporating external knowledge, for example, knowledge graphs have been used to provide additional context for assessing news veracity \cite{yuan2023sustainable}. Graph neural networks that can reason over knowledge graphs have also been applied to fake news detection \cite{phan2023fake}. There is also growing interest in developing interpretable fake news detection models. For example, visualization techniques have been proposed to explain model predictions to end users \cite{szczepanski2021new}.

\textbf{Network Immunization for Misinformation Mitigation}: The discussion on network immunization requires a more comprehensive exploration of existing methods. Proactive immunization strategies \cite{petrescu2021sparse} identify and protect key nodes within social networks to prevent the spread of misinformation. The tree-based approaches \cite{truicua2023mcwdst} like MCWDST prioritize influential nodes based on hierarchical structures. Community-based methods \cite{apostol2023contain} such as ContCommRTD, \cite{apostol2024contcommrtd} target susceptible clusters within the network for localized interventions. Real-time algorithms CONTAIN \cite{apostol2023contain} dynamically detect and mitigate misinformation as it propagates through the network. Incorporating these strategies provide a holistic framework for fake news detection and mitigation.

\textbf{Multimodal Fake News Detection}
There are also multimodal fake news detection methods \cite{tufchi2023comprehensive} that can detect the correlation between image regions and text fragments. Combining information from various modalities, such as textual content \cite{shu2020fakenewsnet}, user metadata \cite{raza2022fake}, and network structures \cite{aimeur_fake_2023}, and using language models \cite{kaliyar_fakebert_2021} can make fake news detection systems more efficient and robust. Our work draws inspiration from prior research, with a particular emphasis on labeling tasks and classification through different methods.
The adoption of multimodal methods \cite{bang2023multitask} emphasizes the necessity to analyze both text and visual content together. Training models to recognize deceptive images is also crucial, as demonstrated by datasets designed for image analysis \cite{heller2018ps,kaliyar_fakebert_2021,yang2023multimodal}.

\begin{table}[ht]
\footnotesize
\centering
\caption{Comparison of our work with state-of-the-art approaches.}
\begin{tabular}{|p{3cm}|p{7cm}|p{1cm}|}
\hline
\textbf{Topic} & \textbf{Key Contributions} & \textbf{Reference} \\
\hline
Multi-platform Fake News Detection&Fake news detection across multiple platforms and languages.  
 Addressed multi-lingual challenges. & \cite{faustini2020fake} \\
\hline
Credibility-based News Verification&Assessed credibility of news articles using ML.  
-Role in fake news detection demonstrated. & \cite{credibilityBased} \\
\hline
Sentiment Analysis Techniques&Emotional tone analysis in text.  
Used ML for falsity inference. & \cite{SentimentAnalysis} \\
\hline
Content-Social Dual Transformer Models&Combined content and social context via dual transformers.  
Enhanced detection precision. & \cite{raza_fake_2022} \\
\hline
Multi-modal Integration in Social Media&Integrated text, images, and publisher features.  
 Improved social media fake news detection. & \cite{evaluatingEffect} \\
\hline
Hybrid BERT-LightGBM Model&Hybridized BERT and LightGBM for classification.  
Improved precision and adaptability. & \cite{HybridBert} \\
\hline
Fuzzy Set Theory in News Analysis&Applied fuzzy set theory for fake news identification.  
Comparative analysis conducted. & \cite{olan2024fake} \\
\hline
Semantic Relationships and Word Embeddings&Captured semantic word relationships.  
Distinguished deceptive content. & \cite{ilie2021context} \\
\hline
Transformers for Fake News Detection&Used transformers (e.g., BERT, GPT) for long text sequences.  
 Enhanced detection accuracy. & \cite{truicua2022misrobaerta} \\
\hline
Sentence and Document Embeddings&Uncovered inconsistencies in sentences.  
 Developed holistic document representations. & \cite{truicua2022awakened,truicua2023s} \\
\hline
Deep Neural Network Ensembles&Combined multiple models to detect diverse characteristics. & \cite{truicua2024danes} \\
\hline
Social Network Immunization&Strengthened key nodes to reduce misinformation spread. & \cite{petrescu2021sparse} \\
\hline
Real-Time Misinformation Detection&Tracked propagation patterns in real-time.  
 Mitigated fake news dynamically. & \cite{jain2016towards} \\
\hline
Community Detection for Network Immunization&Identified susceptible clusters.  
 Designed targeted misinformation interventions. & \cite{zhao2021detecting} \\
\hline
\textbf{Our Work} &  
Dataset, Comparative analysis between  BERT-like vs. LLM .& \textbf{This Study} \\
\hline
\end{tabular}
\label{tab:related}
\end{table}

\textbf{Large Language Models (LLMs) in Detection}
Utilizing natural language processing (NLP) techniques to analyze linguistic features has been a key strategy in improving fake news detection \cite{verma2021welfake}, \cite{qi2019exploiting,gaillard2021countering}. LLMs have emerged as useful methods in the fight against fake news. A related LLM-based fake news detection work \cite{wu2024fake} made use of direct prompting of LLMs with models like GPT-3.5 to provide multi-perspective rationales. This study proposed hybrid approaches to combine the strengths of LLMs and smaller language models to improve detection accuracy. Some works \cite{chen2023combating} have leveraged LLMs to generate style-diverse reframings of news articles, with the goal of enhancing the robustness of fake news detectors against stylistic variations. The potential of online-enabled LLMs like GPT-4, Claude, and Gemini for real-time fake news detection is also explored in a related work \cite{xu2024comparative} to show different results in adapting to emerging misinformation patterns.

Studies have also investigated the use of LLMs in generating explanations and reactions to support misinformation detection \cite{wu2023cheap}. Multimodal approaches combining LLMs with image analysis capabilities have been proposed to tackle cross-domain misinformation \cite{xuan2024lemma,qi2024sniffer}. Despite these advancements, challenges remain. LLMs can sometimes struggle with fact verification tasks, and their performance can vary depending on the specific model and dataset used.

\textbf{Datasets} The development of robust fake news detection systems relies heavily on the diversity and representativeness of the training datasets. It is critical that these datasets encapsulate the complexity of fake news to ensure the models can generalize effectively in real scenarios \cite{raza2023constructing}. Popular benchmarks include FakeNewsNet \cite{shu2020fakenewsnet}, Fakeddit \cite{nakamura2019r}, and NELA-GT \cite{gruppi2020nelagt2019}. Election-related misinformation also continues to get attention, prompting the creation of specialized datasets to tackle misinformation in electoral contexts \cite{allcott2017social,grinberg2019fake,gruppi2020nelagt2019}. In light of the misinformation challenges posed by the COVID-19 pandemic, studies like \cite{alghamdi_towards_2023} have applied transformer models to this new domain of fake news.

\textbf{LLM as Annotator} 
The integration of LLMs for text annotation has been explored through various methodologies, as demonstrated in several recent studies \cite{kim2024meganno+,ni2024afacta,he2023annollm,tan2024large,ozdemir2023quick}. MEGAnno+ \cite{kim2024meganno+} presents a human-LLM collaborative system where LLMs, both proprietary (GPT4) and open-source (Llama, Mistral), initiate the annotation process which is then verified and corrected by humans. AnnoLLM \cite{he2023annollm} is another method that leverages crowdsourcing to enhance the performance of LLMs like GPT-3.5, which shows that carefully designed prompts can significantly improve LLM annotation quality, often matching human-level annotation. One related study on ChatGPT highlights that GPT-3.5 can outperform human crowd-workers in annotation tasks \cite{gilardi2023chatgpt}. These studies collectively suggest that LLMs hold promise for automating annotation processes. In this work, we also utilize LLMs as annotators to label data related to fake news. After initial labeling by the LLMs, we subjected our data to human reviews to ensure reliable labels.

\textbf{Comparison of our work with state-of-the-art work}
Our research contributes to the study of comparing the effectiveness of smaller language models, like those based on BERT, with LLMs in autoregressive settings. We have annotated data using LLMs, but we added the process of human reviews to ensure the validity of our results. Unlike weak supervision or distant labeling that relies on source level news labels, we provided article level labeling to have more reliable ground truth labels. The comparison of our work with related works is briefly depicted in Table \ref{tab:related}

\section{Methodology}
\subsection{Dataset Construction }
\label{data}
\begin{figure}[h]
    \centering
    \includegraphics[width=0.85\linewidth]{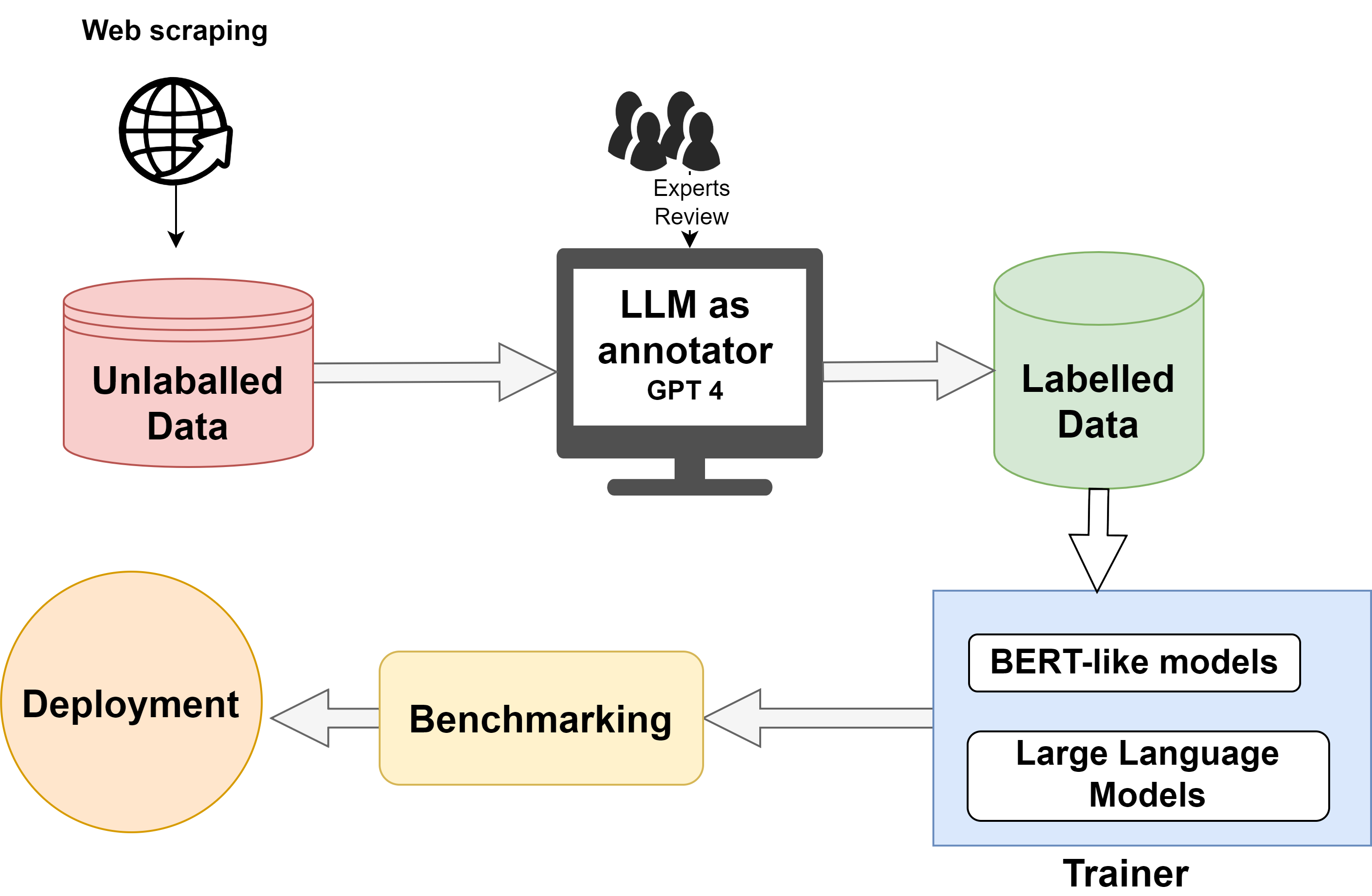}
    \caption{Data Annotation and Model Training Pipeline}
    \label{fig:pipeline}
\end{figure}

We implemented a data annotation and model training pipeline in this work, shown in Figure \ref{fig:pipeline}.  Figure \ref{fig:pipeline} illustrates a workflow for processing and utilizing unlabelled data to create a labeled dataset through LLM based annotations and with human experts' review. Once labeled, the data is stored in a data repository. This labeled dataset is then used to benchmark two families of ML models: BERT-like encoder only models and autoregressive decoder only LLMs. The models are then deployed for practical applications.  We release our annotation method and classification of main models available here \footnote{\href{https://github.com/draip96/FakeNewsClassification}{Code}}.

\subsubsection{Data Source }
We curated about 30,000 news articles using Google RSS based on political news from time period May 2023-Oct 2023.  This portion of the dataset is compiled through Google RSS feeds using the `feedparser` and the `newspaper.Article` API, which facilitate the extraction of article texts, publication dates, news outlets, and URLs. This collection features articles from a diverse array of political news sources, including `The Daily Chronicle', `Global Times News', `Liberty Voice', `New Age Politics', and `The Public Perspective'. We selected a sample of 10,000 news articles, focusing on those with more than 150 words and covering a variety of topics identified during data curation, for labeling using AI. 

\subsubsection{Annotation}
\begin{figure}
    \centering
    \includegraphics[width=0.75\linewidth]{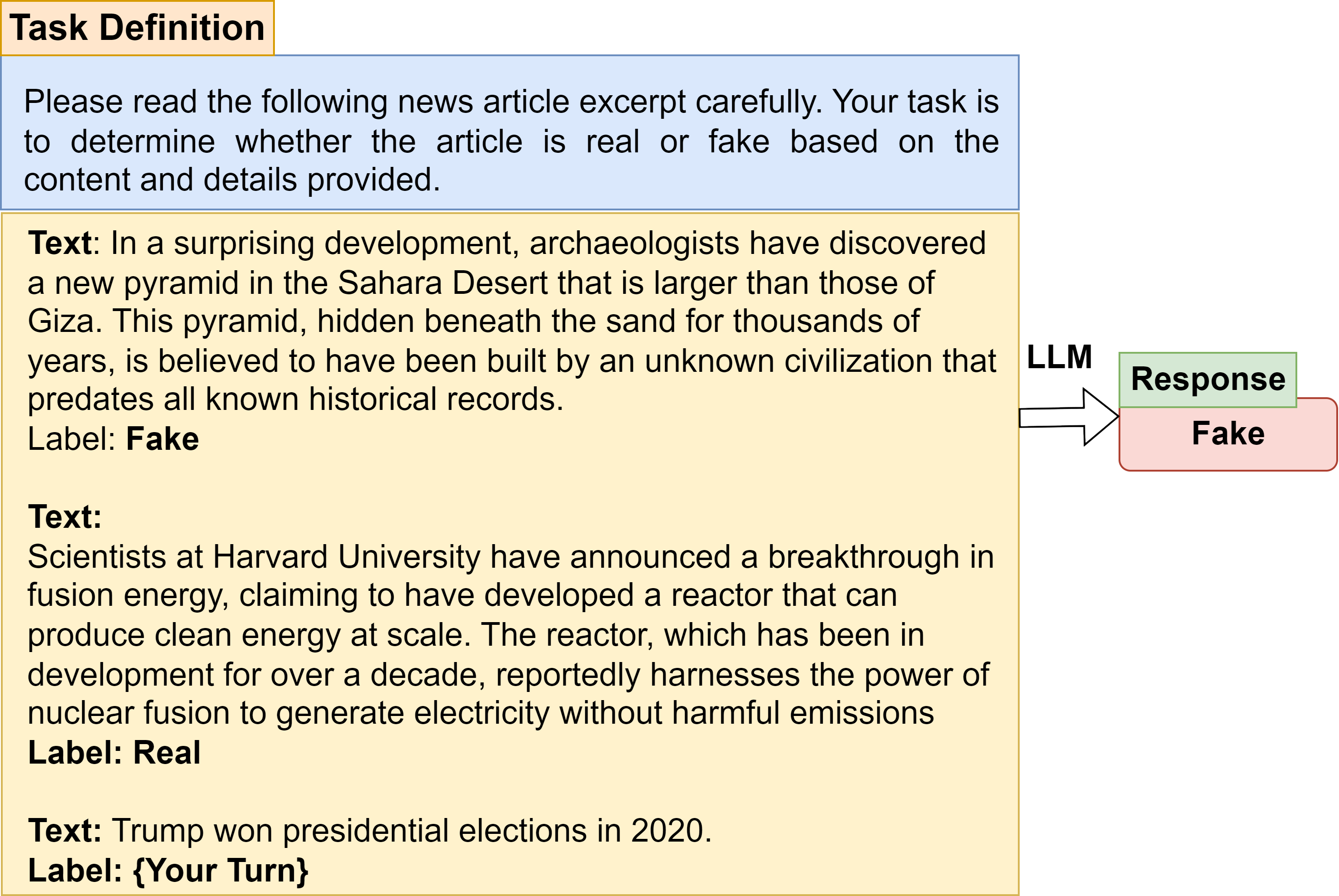}
    \caption{Fake news annotation prompt.}
    \label{fig:prompt}
\end{figure}

To label the selected 10,000 data points curated from Google RSS feeds, we employed the GPT-4 Turbo \cite{openai-gpt4} for binary labeling (fake/real news) annotation. This LLM-based annotation method draws inspiration from recent research works \cite{white2023prompt, gilardi2023chatgpt}, which suggest that annotations from LLMs like ChatGPT are more reliable than those obtained from crowd-sourced workers. In particular, we utilized the GPT-4 Turbo model in a two-shot demonstration mode to assign labels, as illustrated in Figure \ref{fig:prompt}. The decision to focus on a relatively smaller, manageable dataset of 10,000 articles was driven by cost considerations associated with the use of GPT-4 and our aim to achieve highly reliable labels.

\subsubsection{Data Quality Assurance}
\label{quality}
To ensure the accuracy of the annotations generated by the LLM, we implemented a human review system. Over a period of four weeks, each of the 10,000 news articles labeled by the AI was independently reviewed by eight reviewers. The review team comprised of four experts and four students. These reviewers, who come from diverse demographics and disciplines (such as computer science and psychology), were provided with initial guidelines for review. In cases where reviewers disagreed on a label, the article was brought to a team meeting where discrepancies were resolved through consensus to determine the final label. The effectiveness of this collaborative verification process was quantified using an Inter-Rater Reliability Score, which is approximately 72\%. This score reflects a substantial level of agreement among reviewers.

\subsubsection*{Dataset Schema}

The curated dataset consists of 40,000 news articles labeled for fake news detection and the labeled dataset is 10,000. The schema for dataset is given in Table \ref{tab:schema}.
\begin{table}[h]
\centering
\small
\caption{ Schema for the Dataset. Both the unlabeled curated data and labeled data are available upon request.}
\begin{tabular}{|>{\raggedright\arraybackslash}p{0.15\linewidth}|p{0.33\linewidth}|>{\raggedright\arraybackslash}p{0.35\linewidth}|}
\hline
\textbf{Field}            & \textbf{Description}                                                                                  & \textbf{Dataset Part}       \\ \hline
\textbf{Content}          & Full text, avg. 132.8k chars.                                                                         & Available in both unannotated \& annotated \\ \hline
\textbf{Source}           & Origin, avg. 13,857 chars.                                                                            & Unannotated                    \\ \hline
\textbf{Date Published}   & Publication date, 203 unique dates.                                                                   & Unannotated                    \\ \hline
\textbf{Keyword Category} & Articles categorized by keywords, 20 categories.                                                      & Unannotated                    \\ \hline
\textbf{Outlet}           & Media outlet, avg. length not specified.                                                              & Unannotated                    \\ \hline
\textbf{Label}            & GPT-4 Turbo assigns real/fake based on title \& content.& Annotated                      \\ \hline
\end{tabular}

\label{tab:schema}
\end{table}

\subsection{Classification Methods}
We define the task of fake news classification as:
\[
y = f(\mathbf{x}; \theta)
\]
where \( \mathbf{x} \) represents the input features derived from news articles, \( y \) is the binary label indicating fake or real news, and \( \theta \) denotes the parameters of our BERT-like models. The goal is to optimize \( \theta \) such that the prediction \( f(\mathbf{x}) \) closely matches the true label \( y \). 
We have developed a classification system for detecting fake news, integrating models from two model families: BERT-like encoder-only architectures and auto-regressive decoder-only LLMs. 
The goal is to leverage and compare the unique strengths of each model family to enhance the accuracy and efficiency of fake news detection.

\subsubsection{BERT-like Models}
 
An encoder-only model, such as BERT, focuses on understanding the input data by encoding its contextual information without generating new content. These models are particularly useful at understanding context in text, which make them a popular approach for distinguishing between real and fake content. 

Specifically, a BERT-like model processes the input text \( \mathbf{x} \) through a series of transformer layers to produce a contextualized representation \( \mathbf{h} \). This representation is then used to calculate the probability of the news being fake as follows:

\[
\mathbf{h} = \text{BERT}(\mathbf{x}; \theta)
\]

\[
p(\text{fake} | \mathbf{x}; \theta) = \sigma(\mathbf{W}^T \mathbf{h} + b)
\]

where \( \sigma \) is the sigmoid activation function, and \( \mathbf{W} \) and \( b \) are the parameters of a fully connected layer (classification layer) trained to predict the likelihood of the article being fake. The output \( p(\text{fake} | \mathbf{x}; \theta) \) represents the probability that the input \( \mathbf{x} \) is classified as fake news.
In this work, we utilize encoder-only models such as DistilBERT, BERT, and RoBERTa, which are fine-tuned on our dataset.

\subsubsection{Autoregressive LLMs}
Autoregressive decoder-only LLMs, such as the GPT series, Llama, and Mistral models, are trained to predict the next token in a sequence based on the preceding context. For fake news classification, we employ a two-stage approach: fine-tuning and inference with majority voting. The training process begins with data preparation, where we convert our dataset into an instruction format suitable for instruction fine-tuning. We then proceed with instruction fine-tuning, using two-shot demonstrations to provide examples to the model. During training, the model is presented with prompts and correct classifications, allowing it to learn the task of distinguishing between real and fake news articles. The instruction dataset is defined as follows:

\begin{lstlisting} 
<s>[INST] Classify the following news articles as either 'REAL' or 'FAKE':

1. Article: "A recent study published in the Journal of Happiness Research claims that consuming chocolate can significantly increase happiness levels. Researchers surveyed 1,000 participants and found a strong correlation between chocolate consumption and reported happiness".
   Classification: REAL

2. Article: "In an unprecedented event, a group of aliens reportedly landed in Times Square, demanding to meet with President Biden to discuss intergalactic trade. Eyewitnesses claim the aliens looked like giant green octopuses."
   Classification: FAKE

Now classify the following article:

Article: "Breaking News: Scientists Discover Unicorns in the Amazon Rainforest. In a groundbreaking expedition, a team of researchers from the University of Mythical Studies has stumbled upon a herd of unicorns deep in the Amazon rainforest. Dr. Jane Fantasia, lead researcher, stated, 'We were absolutely stunned. These majestic creatures were exactly as described in ancient legends - white horses with spiraling horns on their foreheads'."

[/INST]
\end{lstlisting}

The training configuration involves setting the number of epochs, typically up to 5 for fine-tuning tasks (more epochs led to overfitting), and choosing an appropriate learning rate with a linear decay schedule.
Throughout the training process, we regularly evaluate the model on a validation set to monitor performance and prevent overfitting, tracking the metrics such as accuracy, precision, recall, and F1-score specific to this fake news classification task. We save checkpoints throughout training and select the best-performing model based on the lowest validation loss.

After the training process, we utilize a majority voting mechanism during inference to improve classification reliability and robustness. For each news article, the fine-tuned LLM (e.g., Llama2-7B-Chat or Mistral-7B-Instruct-v0.2) is independently prompted five times with the same input. Each prompt results in a classification ('REAL' or 'FAKE') along with an associated confidence score. The final classification is determined through majority voting, where the most frequently occurring label across the five predictions is selected as the output:

\[
\text{Majority Label} = \text{mode}(\{ \text{label}_1, \text{label}_2, \ldots, \text{label}_n \})
\]

Additionally, we compute the \textbf{average confidence score} from the five responses:

\[
\text{Average Confidence} = \frac{\sum_{i=1}^{n} \text{confidence}_i}{n}
\]

If the average confidence score exceeds a predefined threshold (e.g., 0.8), we adopt the majority label as the final classification. This mechanism addresses the inherent stochasticity of probabilistic LLM outputs by aggregating multiple predictions, thereby reducing the impact of potential outlier responses. Moreover, the ensemble-like effect of majority voting enhances reliability by synthesizing the model probabilistic behavior into a single, consistent prediction.

The choice of five prompts was guided by empirical evaluations, balancing computational efficiency with prediction stability. Similarly, the confidence threshold (0.8) was determined through experiments to ensure robust classification while minimizing false positives and negatives. This method was applied to our instruction fine-tuned versions of Llama2-7B-Chat \cite{Llama} and Mistral-7B-Instruct-v0.2 \cite{jiang2023mistral} for the task of fake news classification. The workflow for classification using Llama2 and Mistral models is outlined below:

\vspace{0.2cm}
\begin{center}
\begin{tikzpicture}[
    node distance=1.8cm and 1.5cm, 
    every node/.style={draw, align=center, minimum height=0.8cm, font=\small, text width=3cm}, 
    scale=0.8, transform shape
]

\node(start) {NEWS ARTICLE};
\node(submit) [right=of start, text width=5cm] {Submit Article to LLM \\ and ask for fake news classification \\ (Multiple Times)};
\node(collect) [right=of submit] {Collect Labels \\ and Confidence Scores};

\node(voting) [below=1.5cm of start] {Majority Voting \\ on Labels};
\node(final) [right=of voting] {Determine \\ FINAL LABEL};
\node(output) [right=of final] {Output \\ FINAL LABEL};

\draw[->] (start) -- (submit);
\draw[->] (submit) -- (collect);

\draw[->] (collect) -- ++(0,-1) -| (voting);

\draw[->] (voting) -- (final);
\draw[->] (final) -- (output);

\end{tikzpicture}
\end{center}
\vspace{0.2cm}








 We fine-tuned these LLMs using parameter-efficient techniques \cite{ding2023parameter} and 4-bit quantization via QLoRA \cite{dettmers2024qlora} to effectively manage GPU memory constraints. While some studies indicate that QLoRA can slightly impact model performance \cite{raschka2023finetuning}, related works on LLM quantization \cite{dettmers2024qlora} suggest that any performance loss due to quantization can be fully recovered through adapter fine-tuning post-quantization. This is the approach we adopted to ensure our models maintained high accuracy and robustness in fake news detection.

\section{Experimental Setting}
\subsection{Evaluation Data}
The dataset developed and used in this study is detailed in Section \ref{data}. It includes 10,000 annotated samples, initially labeled by GPT-4 Turbo and subsequently reviewed by human annotators. This dataset was used for fine-tuning both BERT-like models and LLMs. The data is divided into training, validation, and test sets with an 80-10-10 split ratio.

We also utilized the NELA-GT-2022 dataset \cite{gruppi2023nelagt2022}, a multi-labeled benchmark news corpus designed for studying misinformation, containing 1,778,361 articles from 361 outlets between January 1, 2022, and December 31, 2022. This dataset employs distant supervision for labeling, where article veracity is inferred from the reliability ratings of sources as per the Media Bias/Fact Check (MBFC) database\footnote{\url{https://mediabiasfactcheck.com/}}. We selected a subset of 10,000 articles from NELA-GT-2022, aiming for a balance of fake and real labels and ensuring each article has a minimum length of 50 words to provide substantial content for analysis. For our study, we utilized the news body and source-level label fields provided by the dataset and followed an 80-20 train-test split\footnote{Access to the NELA-GT dataset is available through the dataset's official source or authors, according to their guidelines.}.

To address potential issues arising from data imbalance in our dataset, we implemented the SMOTE (Synthetic Minority Over-sampling Technique) method \cite{chawla2002smote}. This approach increases the frequency of the underrepresented class in the training dataset to match that of the more common class, helping to prevent models from being biased towards the majority class and enhancing their ability to generalize to unseen data. While LLMs are known for their effectiveness at capturing complex patterns even with a limited number of samples \cite{chang2023survey}, their performance can still be affected by imbalanced data. Therefore, we applied the same SMOTE technique for both BERT-like models and LLMs to ensure a fair and consistent experimental setup.

\subsection{Settings and Hyperparameters}
We used a computing cluster equipped with GPUs, including one NVIDIA A100 and four NVIDIA A40s, each with a RAM capacity of 32GB, to ensure efficient data handling and computation. LLMs such as Llama2-7B and Mistral-7B were trained using Parameter-Efficient Fine-Tuning (PEFT) and 4-bit quantization via QLoRA to effectively manage GPU memory constraints.

Our implementation utilizes PyTorch and the Hugging Face Transformers library to efficiently incorporate the BERT encoder layers. For QLoRA \cite{dettmers2024qlora}, we employed the bitsandbytes library along with the Hugging Face Transformers Trainer. We fine-tuned the Llama2-7B-Chat and Mistral-7B-Instruct-v0.2 models on both our dataset and the chosen sample of NELA-GT-2022. The QLoRA configuration included 4-bit NormalFloat (NF4) representation, double quantization, and paged optimizers.

Details of the general hyperparameters and specific QLoRA parameters are provided in Table \ref{tab:hyper-classification}.
\begin{table}[h]
\footnotesize
\centering
\caption{Hyperparameters for fine-tuning BERT-like models and LLM variants for fake news detection.}
\begin{tabular}{p{5.5cm} | p{5.5cm}}
\toprule
\textbf{BERT-like Models} & \textbf{LLM Variants} \\
\midrule
\textbf{Models:} \newline
DistilBERT (base, uncased), BERT (large, uncased), RoBERTa (large) & \textbf{Models:} \newline
Llama2-7B-chat, Mistral-7B-Instruct-v0.2 \\
\textbf{Learning Rate:} \newline
2e-5 to 5e-5 (DistilBERT, BERT), 1e-5 to 4e-5 (RoBERTa) & \textbf{Learning Rate:} \newline
1e-5 to 3e-5 \\
\textbf{Batch Size:} \newline
16-32 & \textbf{Batch Size:} \newline
8 (Training), 16 (Evaluation) \\
\textbf{Epochs:} \newline
20 (early stopping) & \textbf{Epochs:} \newline
5 (early stopping at 1; more led to overfitting) \\
\textbf{Optimizer:} 
AdamW & \textbf{Optimizer:} 
AdamW \\
\textbf{Weight Decay:} \newline
0.01 & \textbf{Weight Decay:} \newline
0.01 \\
\textbf{Warm-up Steps:} \newline
6\% of total steps (RoBERTa only) & \\
\textbf{Classification Threshold:} \newline
0.5 (default for all) & \\
\midrule
\multicolumn{2}{l}{\textbf{QLoRA Parameters}} \\
\multicolumn{2}{>{\raggedright\arraybackslash}p{11cm}}{
\textbf{Parameter:} lora\_r = 64, lora\_alpha = 16, lora\_dropout = 0.2 \newline
\textbf{Task Type:} CAUSAL\_LM \quad \textbf{Bias:} None \newline
\textbf{BitsandBytes:} use\_4bit = True, bnb\_4bit\_dtype = float16, bnb\_4bit\_quant = nf4, use\_nested\_quant = True} \\
\bottomrule
\end{tabular}
\label{tab:hyper-classification}
\end{table}
\begin{table}[h]
\footnotesize
\centering
\caption{Model Performance: Inference Time, Throughput, and Training Time}
\label{tab:model_performance}
\begin{tabular}{>{\raggedright\arraybackslash}p{0.3\textwidth}>{\centering\arraybackslash}p{0.15\textwidth}>{\centering\arraybackslash}p{0.15\textwidth}>{\centering\arraybackslash}p{0.15\textwidth}}
\hline
\textbf{Model}                 & \textbf{Inference Time (ms)} & \textbf{Throughput (examples/sec)} & \textbf{Training Time (hrs)} \\ \hline
BERT\textsubscript{Base-Uncased}              & 5                               & 200.0                              & 0.5                           \\
DistilBERT\textsubscript{Base-Uncased}         & 3                               & 333.3                              & 0.3                           \\
RoBERTa\textsubscript{Base-Uncased}            & 6                               & 166.7                              & 0.7                           \\
Llama2-7B (Zero-Shot)           & 15                              & 66.7                               & -                             \\
Llama2-7B (5-Shots)             & 20                              & 50.0                               & -                             \\
Llama2-7B (Fine-Tuned)          & 12                              & 83.3                               & 3.0                           \\
Mistral-7B-v0.2 (Zero-Shot)     & 14                              & 71.4                               & -                             \\
Mistral-7B-v0.2 (5-Shots)       & 18                              & 55.6                               & -                             \\
Mistral-7B-v0.2 (Fine-Tuned)    & 10                              & 100.0                              & 2.5                           \\ \hline
\end{tabular}
\end{table}

The results were obtained using 5-fold cross-validation, with models trained for up to 20 iterations using early stopping for BERT-like models. For LLMs, we used 5 epochs with early stopping. Our earlier results show that increasing the number of epochs in these LLMs leads to overfitting, as also observed in the Llama-2 paper \cite{Llama}. The mean and standard deviation for each evaluation metric (Accuracy, Precision, Recall, and F1-score) on the test set are reported.
While LLMs typically require more resources, we managed to fine-tune each one in approximately 50 minutes using quantization techniques, with an additional 30 minutes for overall inference. In contrast, BERT-like models completed the process in about minutes in our setup, shown in Table \ref{tab:model_performance}. Running multiple epochs revealed only slight variations.

\subsection{Evaluation Metrics}
We use the standard evaluation metrics used for classification models \cite{truica2017classification}, which are:

\begin{equation}
    Accuracy = \frac{TP + TN}{TP + TN + FP + FN}
\end{equation}

\begin{equation}
Precision = \frac{TP}{TP + FP}
\end{equation}

\begin{equation}
Recall = \frac{TP}{TP + FN}
\end{equation}

\begin{equation}
F1\ score = 2 \times \frac{Precision \times Recall}{Precision + Recall}
\end{equation}
\noindent where TP stands for true positive, TN for true negatives, FN for false positives, and FP for false positives.

For probability-based models like BERT, we use the predicted probabilities directly. A threshold of 0.8 is applied to these probabilities to determine the final classification (REAL or FAKE), allowing us to compute the evaluation metrics.

For LLMs such as Llama2-7B and Mistral-7B, which provide confidence scores rather than probabilities, we adapt our evaluation approach. For each news article, we obtain five classifications with their associated confidence scores. We then use majority voting to determine the final label. The average confidence score is calculated, and if it exceeds our predetermined threshold of 0.8, we accept the majority label as the final classification. All these predictions are compared against the ground truth labels from the testset to compute the final evaluation metrics.
We evaluated the LLMs in two different setups: instruction fine-tuning,  and promting with zero to few shots demonstrations. This comprehensive evaluation allows us to compare the models' performance across different learning paradigms.

\subsection{Exploratory Analysis on our Dataset} 
\begin{figure}[h]
    \centering
    \begin{subfigure}[t]{0.45\textwidth}
        \centering
        \includegraphics[width=\textwidth]{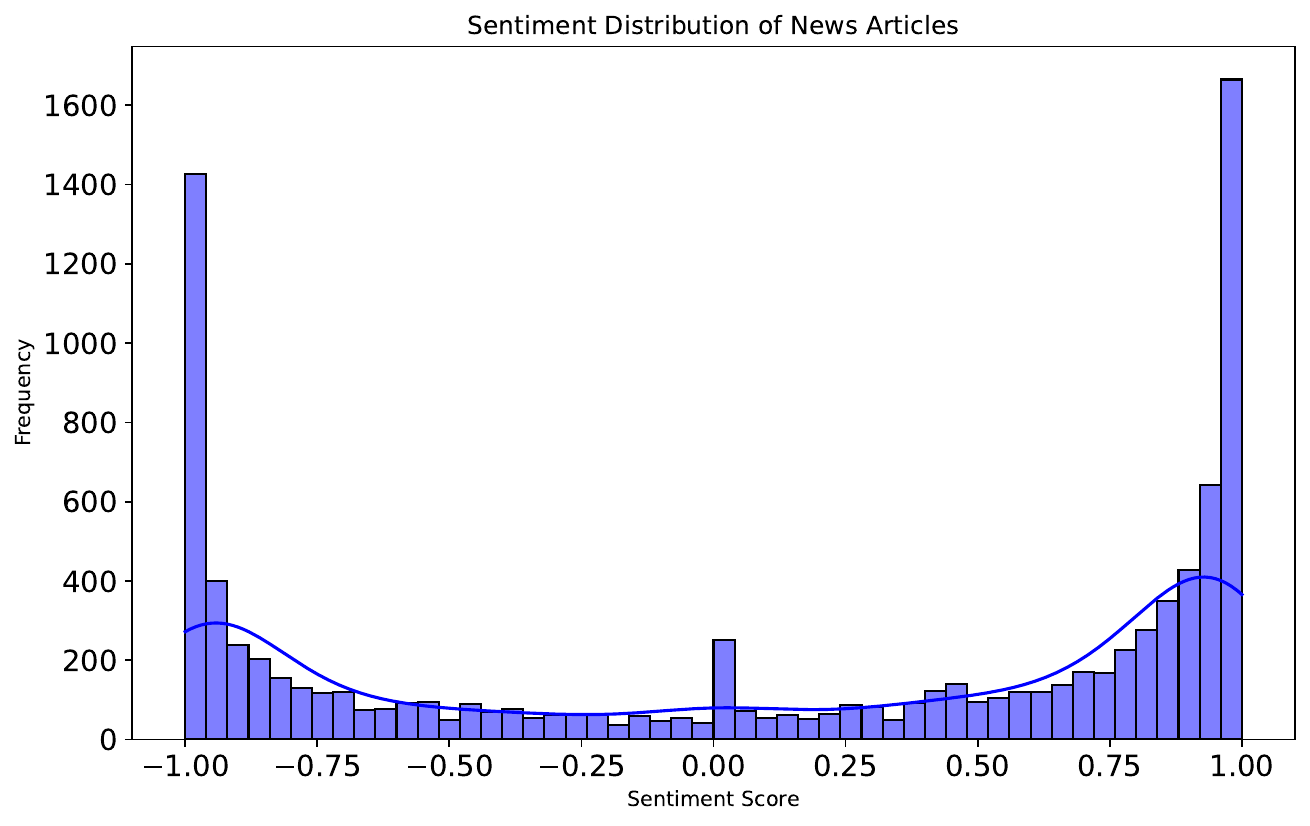}
        \caption{Sentiment Distribution of News Articles}
        \label{fig:sentiment_distribution}
    \end{subfigure}
    \hfill
    \begin{subfigure}[t]{0.45\textwidth}
        \centering
        \includegraphics[width=\textwidth]{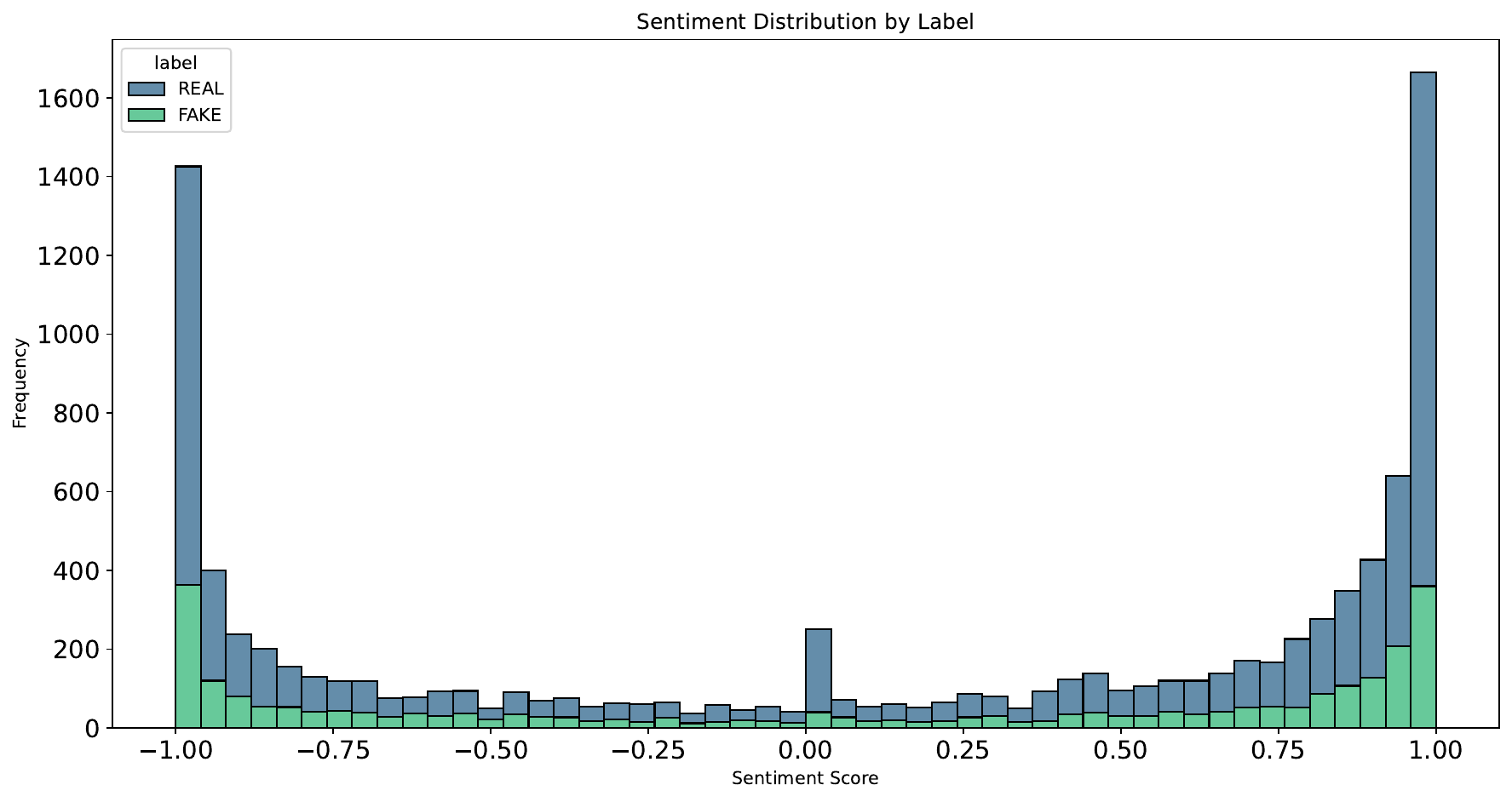}
        \caption{Sentiment Distribution by Label}
        \label{fig:sentiment_distribution_by_label}
    \end{subfigure}
    
    \vspace{-0.3cm}
    
    \begin{subfigure}[t]{0.45\textwidth}
        \centering
        \includegraphics[width=\textwidth]{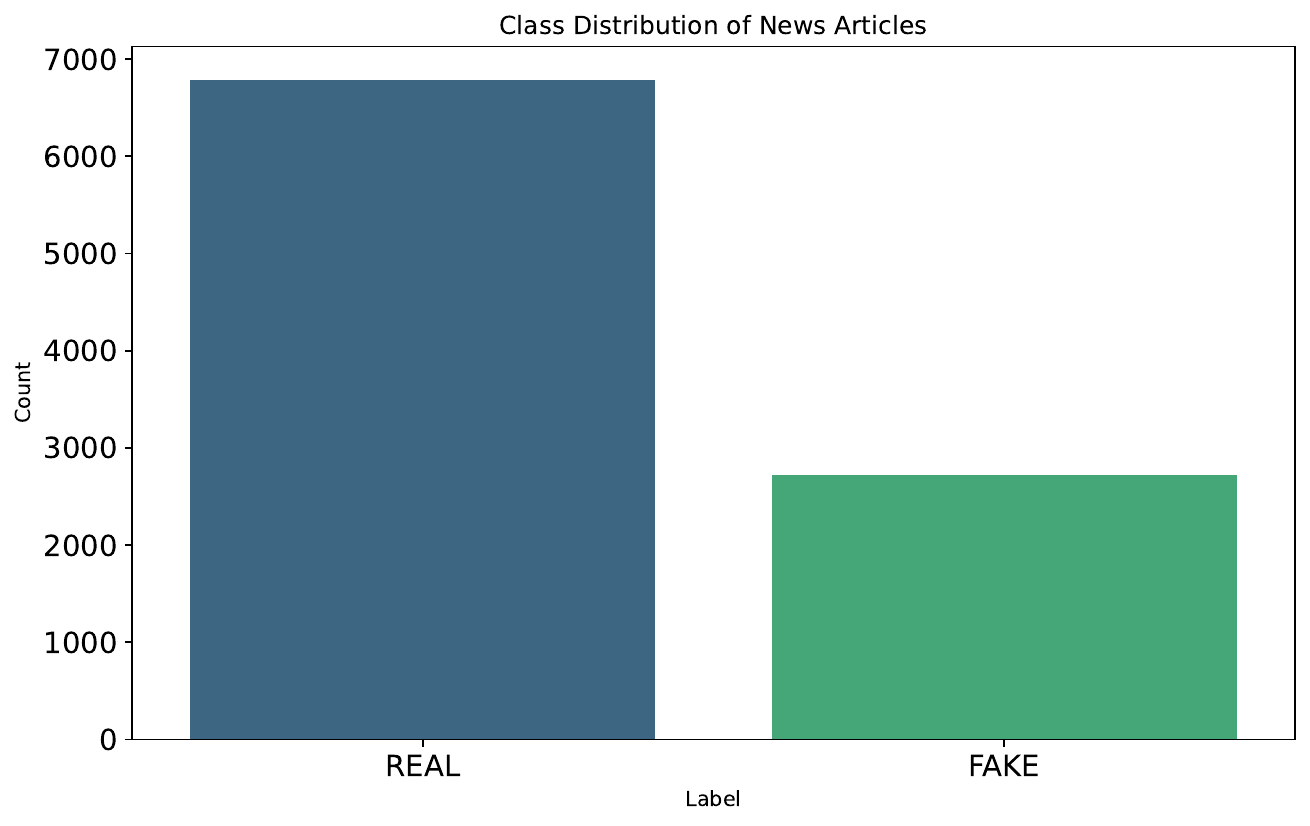}
        \caption{Class Distribution of News Articles}
        \label{fig:class_distribution}
    \end{subfigure}
    \hfill
    \begin{subfigure}[t]{0.45\textwidth}
        \centering
        \includegraphics[width=\textwidth]{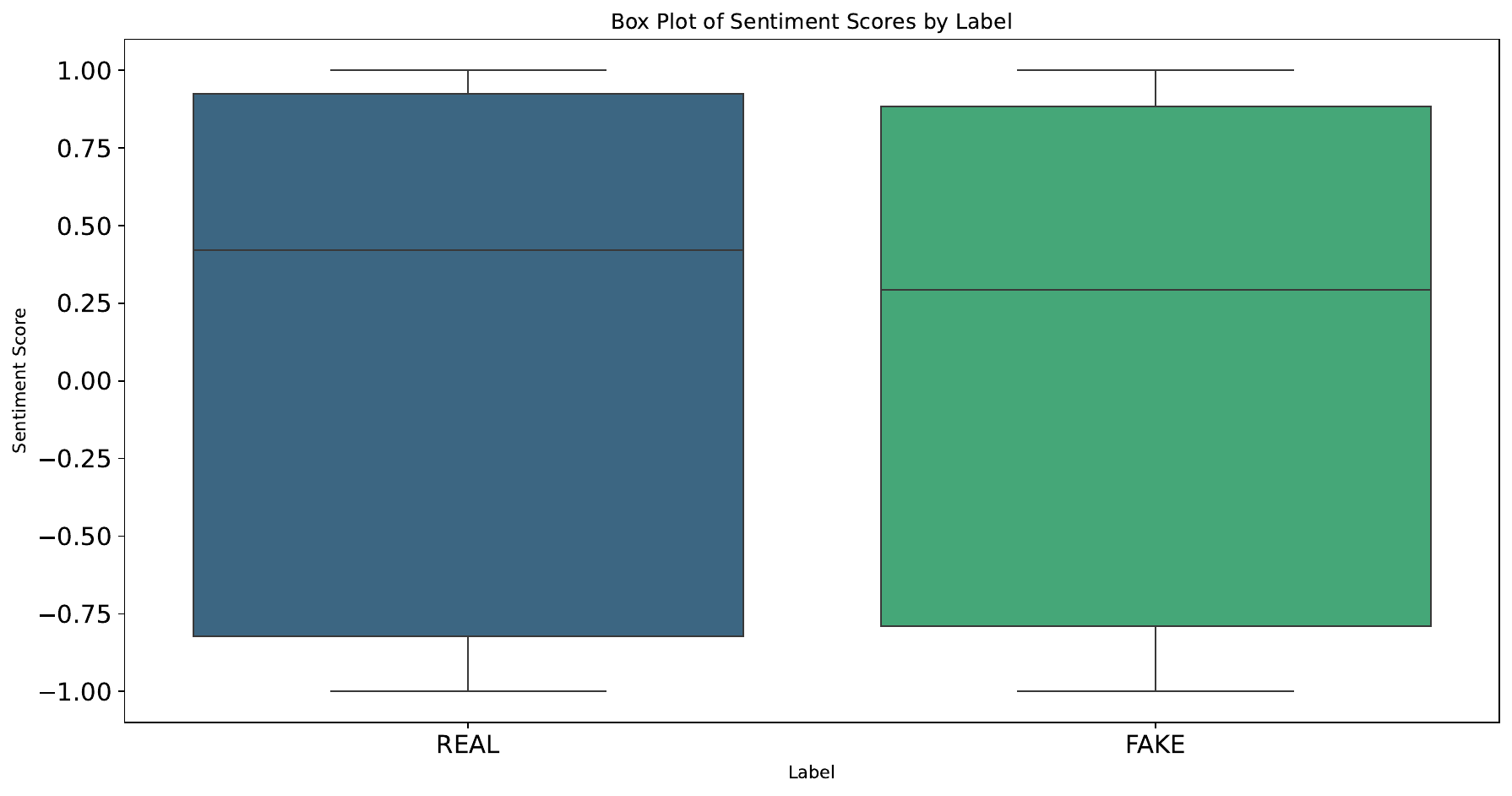}
        \caption{Box Plot of Sentiment Scores by Label}
        \label{fig:box_plot_sentiment_by_label}
    \end{subfigure}
    
    \vspace{-0.3cm}
    
    \begin{subfigure}[t]{0.45\textwidth}
        \centering
        \includegraphics[width=\textwidth]{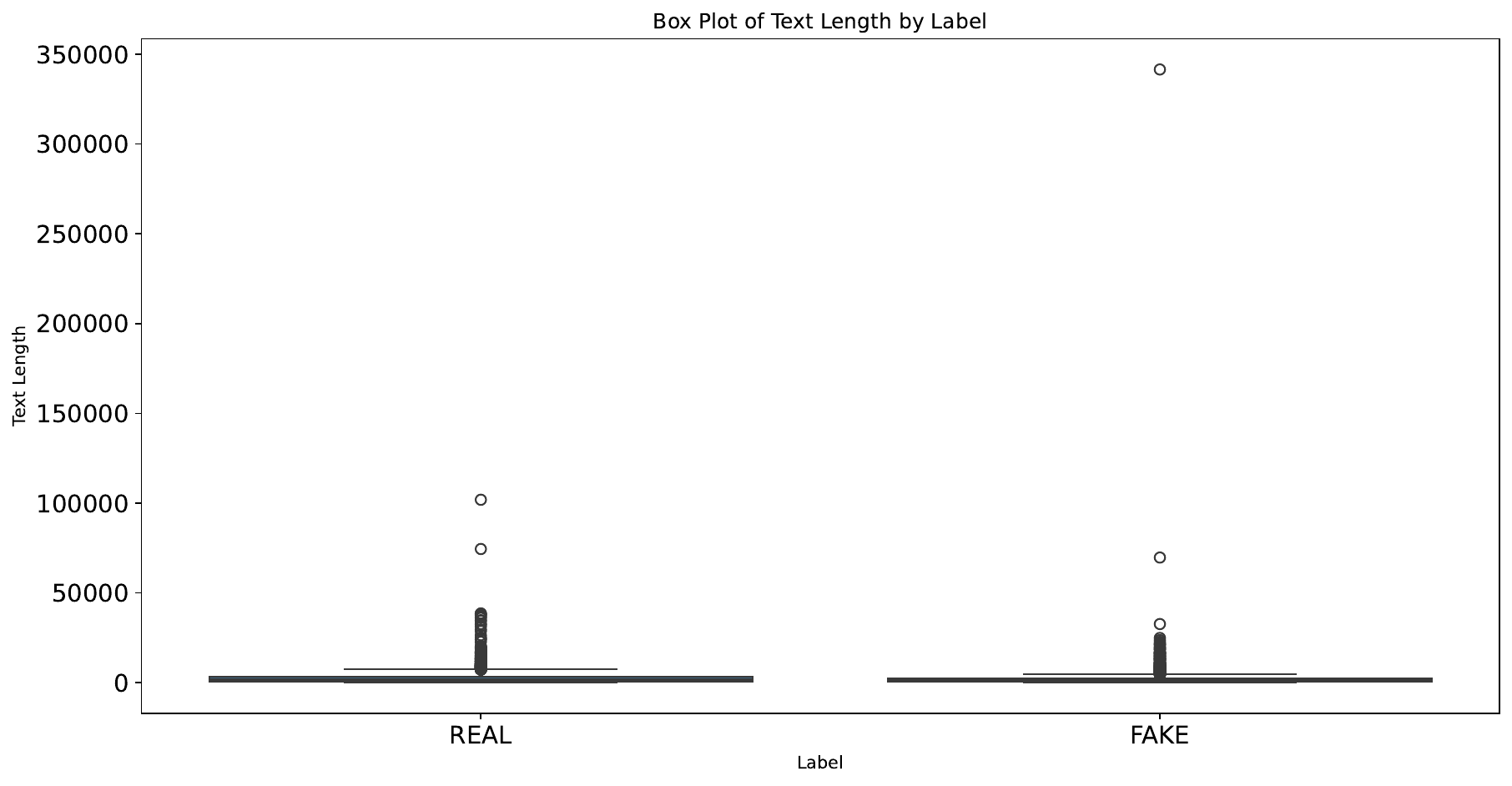}
        \caption{Box Plot of Text Length by Label}
        \label{fig:box_plot_text_length_by_label}
    \end{subfigure}
        \hfill
    \begin{subfigure}[t]{0.45\textwidth}
        \centering
        \includegraphics[width=\textwidth]{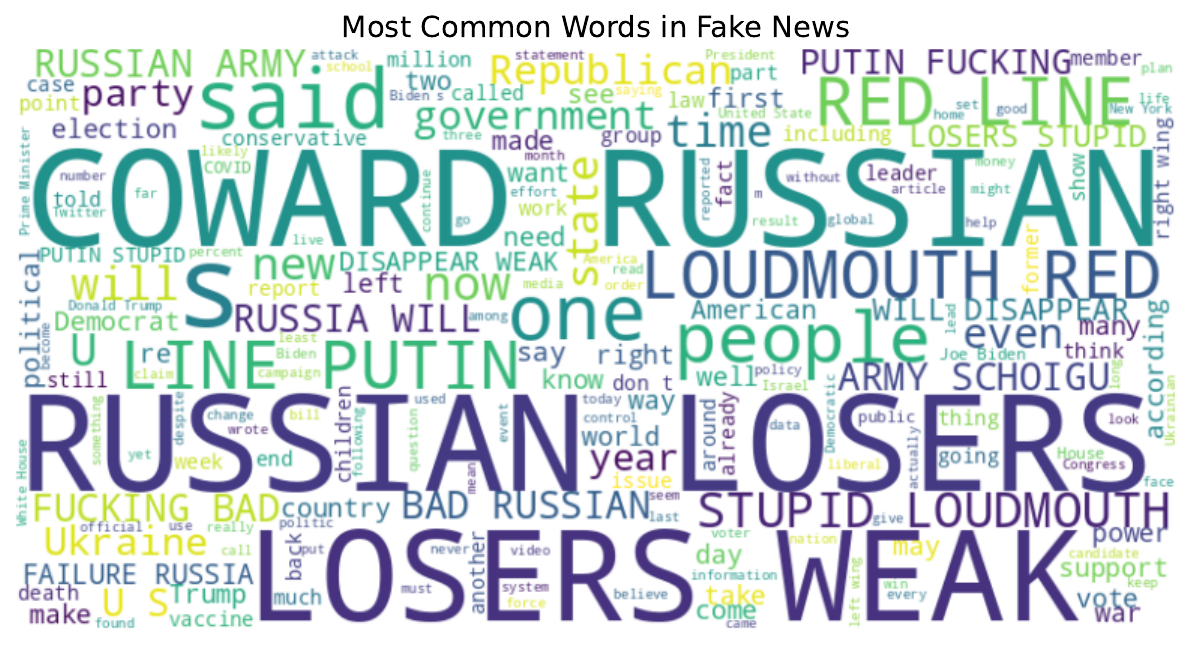}
        \caption{Most Common Words in Fake News}
        \label{fig:word_cloud_fake_news}
    \end{subfigure}

    \caption{Comprehensive Analysis of News Article Sentiment, Distribution, and Content}
    \label{fig:comprehensive_analysis}
\end{figure}

In this section, we present an exploratory analysis on our annotated data (detailed in Section \ref{data}).
Figure \ref{fig:sentiment_distribution} shows a histogram with the distribution of sentiment scores for all news articles. The sentiment scores in our annotated dataset range from -1 (most negative) to 1 (most positive). The distribution is bimodal, with peaks around -1 and 1, indicating that many articles have either very positive or very negative sentiments.

Figure \ref{fig:sentiment_distribution_by_label} shows the stacked histogram that illustrates the sentiment distribution separated by labels (REAL and FAKE). Both real and fake news articles exhibit similar bimodal distributions.
Figure \ref{fig:class_distribution} is a bar plot that shows the count of real and fake news articles in the dataset. 

Figure \ref{fig:box_plot_sentiment_by_label} shows a box plot that compares the sentiment scores for real and fake news articles. Both real and fake news articles have a median sentiment score around 0.5. The interquartile range is also similar, indicating that the sentiment distribution for both labels is quite alike.
Figure \ref{fig:box_plot_text_length_by_label} shows a box plot that shows the distribution of text lengths for real and fake news articles. Real and fake news articles generally have similar text lengths. However, there are some outliers with significantly longer text lengths in both categories.
Figure \ref{fig:word_cloud_fake_news} shows a word cloud that highlights the most frequently used words in fake news articles. Words like ``RUSSIAN", ``LOSERS", ``PUTIN" and ``COWARD" are prominent, suggesting a focus on sensational and provocative language in fake news articles.
We also present main topics derived from our dataset in Table \ref{tab:topics}. 

\begin{table}[ht]
    \footnotesize
    \centering
    \caption{Topical Keyword Distribution from Dataset Analysis}
    \begin{tabular}{|>{\centering\arraybackslash}p{0.2\linewidth}|>{\raggedright\arraybackslash}p{0.75\linewidth}|}
    \hline
    \textbf{Category} & \textbf{Keywords} \\
    \hline
    General Discussions & said, people, just, like, new, time, news, years, health, police \\
    \hline
    Russia and Conflict & russian, weak, losers, coward, putin, stupid, redline, russia, army, bad \\
    \hline
    Global Affairs & said, ukraine, russia, war, china, world, military, biden, energy, united \\
    \hline
    US Politics & trump,president, biden, state, court, republican, election, republicans \\
    \hline
    Political Parties and Elections & party, political, government, election, conservative, democratic, minister, voters, left \\
    \hline
    \end{tabular}
    \label{tab:topics}
\end{table}
Some of our observations on the topics are: 
\textit{Climate Change and Pandemic Response:} Our dataset includes discussions on the environmental impacts of COVID-19 lockdowns. This observation links to studies indicating significant changes in climate patterns during global lockdown periods. \textit{The Information War in Ukraine} We explore the details in the dataset concerning the conflict in Ukraine.  \textit{Developments in Cryptocurrency Markets} The dataset covers topics like fluctuations in the cryptocurrency markets, with an increased focus on central bank digital currencies (CBDCs).  \textit{Policy on Transgender Athletes:} Analysis from our dataset also sheds some light on the sports sector for inclusivity, especially concerning FINA's policies on transgender athletes' participation in competitive swimming. \\

\section{Results and Analysis}

In our study, we aim to address the following research questions (RQ): 
\begin{enumerate} 
\item RQ1: Which families of models, BERT-like models or LLMs, perform better for the task of fake news detection? 
\item RQ2: Are GPT-powered labels more accurate compared to source-level labels? 
\item RQ3:  How do different families of models perform under adversarial perturbations, and which models demonstrate better robustness?
\end{enumerate}

\subsection{Overall Performance}

\begin{table}[htbp]
\centering
\caption{Performance Comparison of Models (in percentages) using 5-fold cross-validation across Precision (P), Recall (R), and F1-score (F1). Evaluations are conducted on our test set with GPT-powered labels and on NELA-GT-2022 with source-level labels. Models include fine-tuned BERT-like models and LLMs. The LLMs used are Llama2-7B-chat (Llama2-7B) and Mistral-7B-v0.2-instruct (Mistral-7B-v0.2), assessed under zero-shot, few-shot (5 examples), and instruction fine-tuning settings. Higher scores ($\uparrow$) indicate better performance, with the best scores highlighted in \textbf{bold}.}
\label{tab:models_performance}
\begin{tabular}{@{}lccc|ccc@{}}
\toprule
\textbf{Model} & \multicolumn{3}{c|}{\textbf{Our Dataset (GPT labels)}} & \multicolumn{3}{c}{\textbf{NELA-GT (Source labels)}} \\
\cmidrule(lr){2-4} \cmidrule(lr){5-7}
 & \textbf{P (\%)} & \textbf{R (\%)} & \textbf{F1 (\%)} & \textbf{P (\%)} & \textbf{R (\%)} & \textbf{F1 (\%)} \\
\midrule
\textbf{BERT-like Models}& & & & & & \\
\hline
BERT\textsubscript{Base-Uncased} & 84.12 & 86.34 & 85.22 & 78.27 & 79.43 & 78.84 \\
DistilBERT\textsubscript{Base-Uncased} & 84.10 & 84.05 & 84.08 & 77.21 & 76.45 & 76.83 \\
RoBERTa\textsubscript{Base-Uncased} & \textbf{89.23} & \textbf{90.14} & \textbf{89.68} & 83.12 & 84.09 & 83.60 \\
\midrule
\textbf{LLMs}& & & & & & \\ \hline
Llama2-7B (Zero-Shot) & 42.15 & 55.37 & 47.75 & 45.27 & 51.48 & 48.16 \\
Llama2-7B (5-Shots) & 53.22 & 58.43 & 55.67 & 49.14 & 52.35 & 50.69 \\
Llama2-7B (Fine-Tuned) & 76.45 & 78.32 & 77.37 & 70.25 & 71.37 & 70.81 \\
Mistral-7B-v0.2 (Zero-Shot) & 53.11 & 57.24 & 55.00 & 45.18 & 48.27 & 46.67 \\
Mistral-7B-v0.2 (5-Shots) & 63.14 & 69.28 & 66.05 & 58.21 & 59.35 & 58.78 \\
Mistral-7B-v0.2 (Fine-Tuned) & 79.33 & 81.14 & 80.23 & 73.12 & 74.19 & 73.65 \\
\bottomrule
\end{tabular}
\end{table}

The main findings from Table \ref{tab:models_performance} indicate that BERT-like models, specifically BERT\textsubscript{Base-Uncased} and RoBERTa\textsubscript{Base-Uncased}, outperform autoregressive LLMs such as Llama2-7B and Mistral-7B-v0.2 across all metrics. RoBERTa, in particular, demonstrates the best performance with a precision of 89.23\% and recall of 90.14\%  on our dataset with GPT-powered labels, which shows the capability of BERT-like models to provide accurate and reliable predictions in such classification tasks. The BERT and DistilBERT models also show strong performance, and offer a balance between accuracy and computational efficiency. This makes these encoder only models suitable for resource-constrained environments for the classification task.

The use of GPT-powered labeling with human supervision proves to be more effective in detecting fake news than source-level distant labeling. This suggests that human insights significantly enhance the labeling process. On the NELA-GT-2022 dataset with source-level labels, the performance of all models slightly decreases, which indicates that the GPT-powered labels are generally more accurate.

Within the LLMs, instruction fine-tuning shows the best results, followed by few-shot learning, and then zero-shot learning. This result shows that task-specific fine-tuning can perform better and that more data points are good than just prompting. However, some underperformance of LLMs compared to BERT-like models suggests that the generative nature of LLMs, which are often trained for text generation rather than classification. 

\textit{Discussion}: Overall, the results demonstrate that GPT-powered labels offer superior accuracy, and that the BERT-like fine-tuned models are quite effective classifiers for this task of fake news detection. The results answer that for the task of fake news detection, LLMs outperform BERT-like models (RQ1), and GPT-powered labels demonstrate higher accuracy compared to source-level labels (RQ2).

Given the better performance of our dataset with GPT-powered labels (over weak labels by NELA-GT dataset), further results will focus on this dataset.

\subsection{Ablation Study of BERT-like Models}
The ablation study in Table \ref{table:ablation_study} examines the performance of BERT, RoBERTa, and DistilBERT models under varying batch sizes, learning rates, and training epochs with early stopping. This analysis focuses on encoder-only models, known for their effectiveness in classification tasks, to understand the impact of hyperparameters. LLMs, being autoregressive and designed for generative tasks, are excluded from this study due to their distinct training and evaluation requirements.

\begin{table}[h]
\centering
\setlength{\tabcolsep}{6pt} 
\caption{Performance of BERT, RoBERTa, and DistilBERT models with varying batch sizes and learning rates (LR). Results are shown for Accuracy (Acc) and F1-score (F1). Best model scores are in \textbf{bold}, second-best \textit{italic}, and third-best as \underline{underline}.}
\label{table:ablation_study}
\begin{tabular}{@{}p{1cm}p{1cm}p{0.9cm}p{0.9cm}p{0.9cm}p{0.9cm}p{0.9cm}p{0.9cm}@{}}
\toprule
\textbf{Batch} & \textbf{LR} & \multicolumn{2}{c}{\textbf{BERT}} & \multicolumn{2}{c}{\textbf{RoBERTa}} & \multicolumn{2}{c}{\textbf{Distil-BERT}} \\ 
\cmidrule(lr){3-4} \cmidrule(lr){5-6} \cmidrule(lr){7-8}
\textbf{Size} &  & \textbf{Acc} & \textbf{F1} & \textbf{Acc} & \textbf{F1} & \textbf{Acc} & \textbf{F1} \\
\midrule
\multirow{3}{*}{8}  & 5e-5 & 81.24 & 81.19 & 82.37 & 82.45 & 80.12 & 80.34 \\
                    & 3e-5 & 84.49 & 83.27 & 85.31 & 84.22 & 83.29 & 82.15 \\
                    & 2e-5 & 82.03 & 82.14 & 83.22 & 83.33 & 81.48 & 81.29 \\
\midrule
\multirow{3}{*}{16} & 5e-5 & 83.15 & 83.08 & 84.27 & 84.31 & 82.29 & 82.45 \\
                    & 3e-5 & \textit{88.16} & \textit{85.38} & \textbf{92.12} & \textbf{89.47} & \underline{85.19} & \underline{84.21} \\
                    & 2e-5 & 83.22 & 83.11 & 84.34 & 84.25 & 82.01 & 82.09 \\
\midrule
\multirow{3}{*}{32} & 5e-5 & 80.18 & 80.23 & 81.41 & 81.29 & 79.14 & 79.22 \\
                    & 3e-5 & 83.11 & 83.02 & 84.19 & 84.11 & 82.33 & 82.25 \\
                    & 2e-5 & 81.12 & 81.18 & 82.24 & 82.35 & 80.22 & 80.14 \\
\bottomrule
\end{tabular}
\end{table}

Table \ref{table:ablation_study} presents an ablation study demonstrating that BERT-like models achieve optimal performance with a batch size of 16 and a learning rate of 3e-5 in this setup. This configuration yields the highest accuracy and F1-scores, particularly for RoBERTa, which achieves 92.12\% accuracy and 89.47\% F1-score. 

\textit{Discussion}: The study highlights the importance of optimizing batch size, learning rate, and other hyperparameters to enhance the performance of BERT-like models in classification tasks.

\subsection{Analysis of Few-Shot Learning Versus Fine-Tuning with Large Language Models}
\begin{figure}[ht]
    \centering
    \begin{subfigure}[b]{\textwidth}
        \centering
        \includegraphics[width=0.75\textwidth]{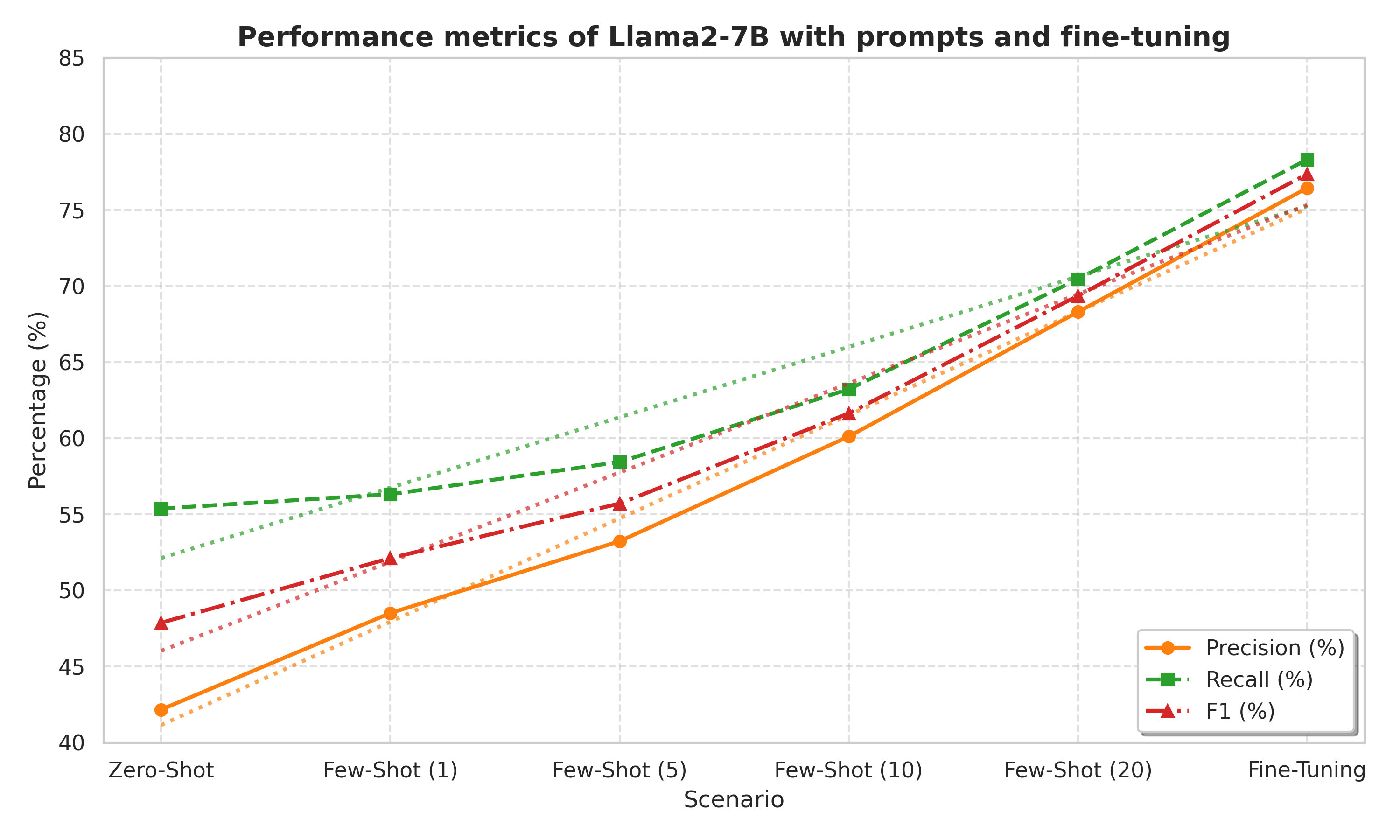}
        \caption{Performance of Llama2-7B}
        \label{fig:Llama2-7b}
    \end{subfigure}
    

    \begin{subfigure}[b]{\textwidth}
        \centering
        \includegraphics[width=0.75\textwidth]{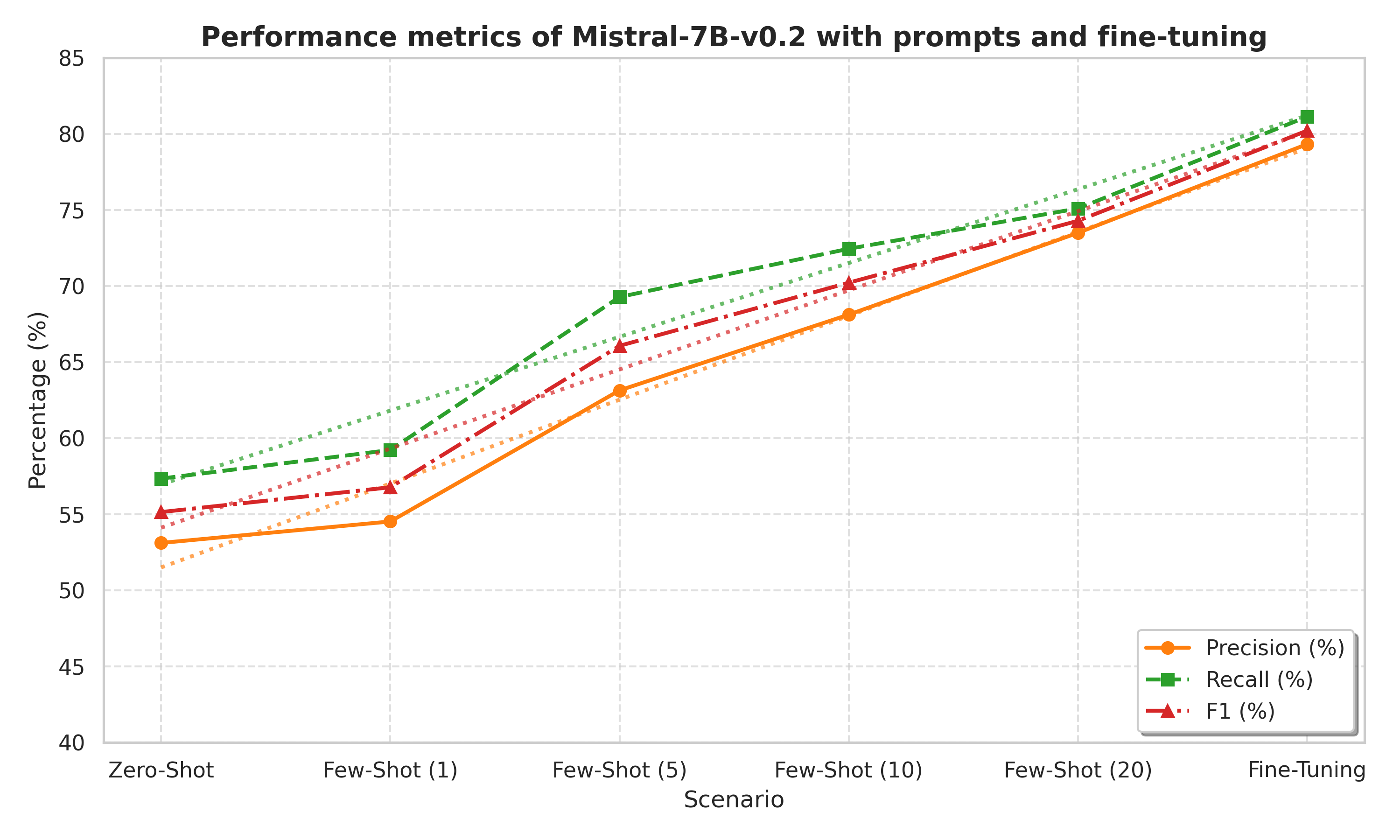}
        \caption{Performance Metrics of Mistral-7B-v0.2}
        \label{fig:mistral-7b-v0.2}
    \end{subfigure}
  \caption{Comparative analysis of performance metrics for Llama2-7B and Mistral-7B-Instruct-v0.2 across various prompt settings and fine-tuning ccenarios}

    \label{fig:performance-comparison}
\end{figure}

To evaluate the performance of LLMs, we conducted experiments across zero-shot, few-shot, and fine-tuning scenarios. These experiments help illustrate how well the models can generalize with varying amounts of task-specific information and training data. The results are shown in Figure \ref{fig:performance-comparison}.

The results in Table \ref{fig:performance-comparison} demonstrate that increasing the number of demonstrations in prompts leads to better performance for both Llama2-7B-chat and Mistral-7B-v0.2-instruct. For instance, 10 demonstrations per class outperform 5 demonstrations, with this trend continuing as more examples are added. While these models show the ability to perform reasonably well with limited examples, fine-tuning with task-specific data shows the best results for both models.

\textit{Discussion}: The results indicate that fine-tuning is a superior approach for classification tasks compared to few-shot demonstrations, although both methods show improvements over zero-shot performance.

\subsection{Impact of Text Perturbations on Model Predictions}
To assess the robustness of various fake news detection models, we conducted experiments using the TextAttack\footnote{\href{https://textattack.readthedocs.io/en/master/}{TextAttack Library} }  library to apply several text perturbations. Our goal was to determine how minor textual modifications affect the predictive accuracy of BERT, RoBERTa, DistilBERT, Llama2-7B, and Mistral-7B models. These tests simulated real-world scenarios where news titles might have subtle but significant variations. Table \ref{tab:model_impact} presents key examples of our findings.

The results in Table \ref{tab:model_impact} reveal significant variations in model responses to text perturbations. For example, `Drop' and `Insertion' perturbations caused some models to misclassify real news as fake, while `Swap' and `Rephrase' perturbations generally maintained high accuracy across models, indicating robustness against straightforward syntactic changes.

BERT-like models demonstrated high accuracy, particularly with `Swap' and `Rephrase' perturbations. However, they showed vulnerabilities to `Drop' and `Insertion' perturbations, likely due to their reliance on specific word patterns learned during training. In contrast, LLMs (Llama2-7B and Mistral-7B) showed better consistency across all perturbation types, maintaining high accuracy even with challenging perturbations like `Drop' and `Insertion'. This robustness can be attributed to their extensive pre-training on large, diverse datasets, equipping them to handle a variety of text manipulations effectively.

\textit{Discussion}: Our findings indicate that LLMs are more adept at handling text perturbations in fake news classification tasks compared to BERT-like models. This superior performance likely stems from their advanced contextual understanding and generative capabilities, allowing them to adapt more effectively to data variations. The result answers our RQ3 and reveals that LLMs exhibit superior performance and robustness under adversarial perturbations compared to other model families.
{\footnotesize
\begin{longtable}{|p{1cm}|p{2.2cm}|p{1cm}|p{1cm}|p{1.3cm}|p{1cm}|p{1cm}|p{1cm}|}
\caption{Impact of Text Perturbations on Fake News Detection Models} \label{tab:model_impact} \\
\hline
\textbf{Perturb-ation} & \textbf{Example Text} & \textbf{Original Label} & \textbf{BERT} & \textbf{RoBERTa} & \textbf{Distil BERT} & \textbf{Llama2-7B} & \textbf{Mistral-7B} \\ \hline
\endfirsthead
\caption[]{(continued)} \\
\hline
\textbf{Perturb-ation} & \textbf{Example Text} & \textbf{Original Label} & \textbf{BERT} & \textbf{RoBERTa} & \textbf{Distil BERT} & \textbf{Llama2-7B} & \textbf{Mistral-7B} \\ \hline
\endhead
\hline \multicolumn{8}{|r|}{{Continued on next page}} \\ \hline
\endfoot
\hline
\endlastfoot
Original & CDC confirms new treatment effective against flu & Real & Real & Real & Real & Real & Real \\ \hline
Swap & New treatment confirmed effective against flu by CDC & Real & Real & Real & Real & Real & Real \\ \hline
Drop & New treatment effective against flu & Real & Fake & Fake & Fake & Real & Real \\ \hline
Insertion & CDC rumor: new treatment might be effective against flu & Real & Fake & Fake & Real & Real & Real \\ \hline
Rephrase & CDC states treatment can combat flu effectively & Real & Real & Real & Real & Real & Real \\ \hline
Negation & CDC confirms new treatment not effective against flu & Real & Fake & Fake & Fake & Real & Real \\ \hline
Synonym & CDC verifies new remedy effective against flu & Real & Real & Real & Real & Real & Real \\ \hline
Antonym & CDC denies new treatment effective against flu & Real & Fake & Fake & Fake & Real & Real \\ \hline
Typo & CDC confirms nuw treatment effective against flu & Real & Fake & Fake & Fake & Real & Real \\ \hline
Paraphrase & CDC announces effectiveness of new flu treatment & Real & Real & Real & Real & Real & Real \\ \hline
Shuffle & Treatment new CDC confirms effective against flu & Real & Fake & Fake & Fake & Real & Real \\ \hline
Punctuation & CDC confirms new treatment: effective against flu & Real & Real & Real & Real & Real & Real \\ \hline
Case & cdc confirms new treatment effective against flu & Real & Real & Real & Real & Real & Real \\ \hline
Contraction & CDC's new treatment effective against flu & Real & Real & Real & Real & Real & Real \\ \hline
Emphasis & CDC confirms \textbf{new} treatment effective against flu & Real & Real & Real & Real & Real & Real \\ \hline
\textbf{Correct per Model} & 15 examples & & 10 & 10 & 10 & 14 & 14 \\ \hline
\end{longtable}
}

\section{Discussion}

Our study is aimed to determine whether BERT-like models or LLMs enhance fake news detection and how their advantages can be effectively utilized for this task. The results indicate that while LLMs such as Llama and Mistral offer valuable tools for annotation and provide insightful analyses, BERT-like models generally outperform LLMs in classification tasks. However, LLMs perform better in handling perturbations, they demonstrate robust performance under varying conditions. 

These findings suggest that  BERT-like models, despite not as robust as LLMs can be good classifiers, and can complement them when cost and compute is a challenge. However, LLMs can provide much resilience against data perturbations and attacks. This dual approach leverages the strengths of both model types: the precision of BERT-like models in stable environments \cite{raza2024nbias} and the adaptability of LLMs in different scenarios. 

\subsection*{Limitations and Future Directions}

Like any research endeavor, this study has some limitations too, which are discussed below:

 As we employ quantization techniques, such as 4-bit quantization via QLoRA, to manage GPU memory constraints in our autoregressive LLMs, there may be some performance degradation compared to their full-precision BERT-like models counterparts. This might result in slightly lower accuracy or reliability in some instances, but these reductions are typically nominal, as discussed in the state-of-the-art also \cite{ranasinghe_mudes_2021}.
    
    Conversely, BERT-like models in our setup are utilized without such memory management techniques, which allow these models to operate at full capacity. The results suggests that BERT-like models may demonstrate high performance in certain contexts, it does not necessarily mean they will outperform LLMs in all scenarios. Our goal in this study just to provide a balanced comparison, where we highlight where each model family performs better and where it may face limitations.
   
     Our study did not evaluate other well-known LLMs, such as Claude or GPT-3.4/4, for fake news classification due to two main reasons: firstly, we utilized GPT-4 Turbo as an annotator, which precluded its use as a classifier in our experiments; secondly, the weights for these models are not publicly accessible. Our aim was to focus on classifiers that are both open-source and freely available, such as Llama and Mistral.
 
     Our study is limited to analyzing perspectives derived solely from the LLMs' responses. We did not explore other prompting techniques, such as chain-of-thought methods, which may give different insights. Additionally, while we utilized the instruction versions of Llama and Mistral models, their non-instruction versions, which are capable of sequence classification, were not examined in this research.
    
    While our annotation method shows promise, it currently has limitations for production-level annotation. One potential enhancement could involve integrating active learning \cite{settles2009active}, a ML approach where the model iteratively queries a human annotator \cite{settles2009active} or utilizes an LLM as an annotator \cite{rouzegar2024enhancing}. This method focuses on labeling new data points that the model deems most informative, thereby iteratively and efficiently improving its performance. These limitations suggest that there is significant potential for enhancing the operational effectiveness of LLMs.
 
     In our evaluation of LLMs, the confidence scores provided by LLMs, which are often treated as continuous values indicative of model certainty, differ from the discrete probability scores produced by BERT-like models. This discrepancy can lead to challenges in interpreting and integrating the outputs of these models within traditional ML frameworks.
       We have not explored potential of LLMs in multilingual and multimodal contexts, which could significantly expand their applicability in global fake news detection scenarios.

Overall, our findings highlight the need for continued exploration of BERT-like models and LLM capabilities and the development of strategies to effectively integrate these models with task-specific models. 

 \section{Conclusion}
This study investigates the capabilities of BERT-like models and LLMs in the detection of fake news, using a novel dataset annotated by GPT-4-turbo and verified by human reviewers. Our findings indicate that BERT-like models, in general, perform better in classification tasks due to their precision and computational efficiency. However, LLMs  demonstrate resilience in handling text perturbations, which is attributed to their advanced contextual understanding and generative capabilities. The integration of gen. AI for annotation, combined with human oversight, proves to be a robust approach for enhancing the accuracy of fake news detection. Future research should explore optimizing prompting techniques for LLMs and integrating multimodal data to further improve fake news detection system. 


\section*{Declarations}

\textbf{Funding:} The second author was supported by NSERC University Student Research Award and the third author is supported by NSERC (Grant 2020-04760) 

\textbf{Conflict of interest/Competing interests:} The authors declare no conflict of interest.

\textbf{Ethics approval and consent to participate: }Research Ethics Board approval is not required because no human participants were involved in this work.

\textbf{Consent for publication: }Not applicable.

\textbf{Data availability: }The data used in this work can be obtained from the first author.

\textbf{Code availability: }The code used in this work is made available at \href{https://github.com/draip96/FakeNewsClassification}{Code}.

\textbf{Author contributions}: S.R. conceptualized the problem and work plan.  S.R. and D.P. annotated the data. D.P. prepared the data and performed BERT baselines. S.R. prepared LLM baselines. C.D. reviewed the results. S.R. wrote the first version and revised, and all authors reviewed it.

\bibliography{sn-bibliography}

\end{document}